\pdfoutput=1

\documentclass[11pt]{article}

\usepackage[]{emnlp2021}

\usepackage{times}
\usepackage{latexsym}

\usepackage[T1]{fontenc}

\usepackage[utf8]{inputenc}

\usepackage{microtype}
\usepackage{times}
\usepackage{latexsym}
\usepackage{amsmath,graphicx}
\usepackage{latexsym}
\usepackage{algorithm}
\usepackage{algpseudocode}
\usepackage{url}
\usepackage{multirow}
\usepackage{CJKutf8} 
\usepackage{textcomp }
\usepackage{comment} 
\usepackage{circledsteps}

\usepackage{amsmath,graphicx}
\usepackage{latexsym}
\usepackage{algorithm}
\usepackage{algpseudocode}
\usepackage{url}
\usepackage{multirow}
\usepackage{CJKutf8}
\usepackage{textcomp}
\usepackage{comment}
\usepackage[notextcomp]{stix}
\usepackage{xspace}
\usepackage{listings}
\usepackage{booktabs}
\usepackage{adjustbox}
\usepackage{enumitem}
\usepackage{appendix}
\usepackage{cleveref}

\usepackage{todonotes}

\newcommand{\mari}[2][]{\todo[color=cyan,size=\scriptsize,fancyline,caption={},#1]{Mari:#2}}

\title{Dialogue State Tracking with a Language Model\\ using Schema-Driven Prompting}

 \author{Chia-Hsuan Lee \\
 University of Washington\\
   \texttt{chiahlee@uw.edu} \\\And
   Hao Cheng \\
   Microsoft Research \\
   \texttt{chehao@microsoft.com}\\\And
   Mari Ostendorf\\
   University of Washington\\
   \texttt{ostendor@uw.edu
}}

\begin{document}
\maketitle
\begin{abstract}
Task-oriented conversational systems often use dialogue state tracking to represent the user's intentions, which involves filling in values of pre-defined slots. Many approaches have been proposed, often using task-specific architectures with special-purpose classifiers.  Recently, good results have been obtained using more general architectures based on pretrained language models.
Here, we introduce a new variation of the language modeling approach that uses schema-driven prompting to provide task-aware history encoding that is used for both categorical and non-categorical slots. 
We further improve performance by augmenting the prompting with schema descriptions, a naturally occurring source of in-domain knowledge.
Our purely generative system achieves state-of-the-art performance on MultiWOZ 2.2 and achieves competitive performance on two other benchmarks: MultiWOZ 2.1 and M2M. The data and code will be available at \url{https://github.com/chiahsuan156/DST-as-Prompting}.
\end{abstract}


\section{Introduction}

In task-oriented dialogues, systems communicate with users through natural language to accomplish a wide range of tasks, such as food ordering, tech support, restaurant/hotel/travel booking, etc.
The backbone module of a typical 
system is dialogue state tracking (DST), where the user goal is inferred from the dialogue history \cite{henderson2014second,shah2018building,budzianowski2018multiwoz}.
User goals are represented in terms of values of pre-defined slots associated with a schema determined by the information needed to execute task-specific queries to the backend.
In other words, user goals are extracted progressively via slot filling based on the schema throughout the conversation.
In this paper, we focus on multi-domain DST where the dialogue state is encoded as a list of triplets in the form of (\textit{domain, slot, value}), e.g. (\textit{``restaurant'', ``area'', ``centre''}). 

There are two broad paradigms of DST models, \textit{classification-based} and \textit{generation-based} models, where the major difference is how the slot value is inferred.
In classification-based models \cite{ye2021slot,chen2020schema}, the prediction of a slot value is restricted to a fixed set for each slot, and non-categorical slots are constrained to values observed in the training data.
In contrast, generation-based models \cite{wu2019transferable,kim2020efficient} decode slot values sequentially (token by token) based on the dialogue context, with the potential of recovering unseen values.
Recently, generation-based DST built on large-scale pretrained neural language models (LM) achieve strong results without relying on domain-specific modules.
Among them, the autoregressive model \cite{peng2020soloist,hosseini2020simple} uses a uni-directional encoder whereas the sequence-to-sequence model \cite{lin-etal-2020-mintl,heck2020trippy} represents the dialogue context using a bi-directional encoder.


In this study, we follow a generation-based DST approach using a pre-trained sequence-to-sequence model, but with the new strategy of adding task-specific prompts as input for sequence-to-sequence DST models, inspired by \textit{prompt-based} fine-tuning \cite{radford2019language,gpt3}.
Specifically, instead of generating domain and slot symbols in the decoder, 
we concatenate the dialogue context with domain and slot prompts as input to the encoder, where prompts are taken directly from the schema.
We hypothesize that jointly encoding dialogue context and schema-specific textual information can further benefit a sequence-to-sequence DST model.
This allows task-aware contextualization for more effectively guiding the decoder to generate slot values.

Although the domain and slot names typically have interpretable components, they often do not reflect standard written English, e.g.\ ``\textit{arriveby}'' and ``\textit{ref}''.
Those custom meaning representations are typically abbreviated and/or under-specified, which creates a barrier for effectively utilizing the pretrained LMs.
To address this issue, we further incorporate natural language schema descriptions into prompting for DST,
which include useful information to guide the decoder.
For example, the description of ``\textit{ref}'' is ``\textit{reference number of the hotel booking}'';
the values of ``\textit{has\_internet}'' are ``\textit{yes}'', ``\textit{no}'', ``\textit{free}'', and ``\textit{don't care}''.

In short, this work advances generation-based DST in two ways.  First, candidate schema labels are jointly encoded with the dialogue context, providing a task-aware contextualization for initializing the decoder.  Second, natural language descriptions of schema categories associated with database documentation are incorporated in encoding as prompts to the language model, allowing uniform handling of categorical and non-categorical slots. When implemented using a strong pretrained text-to-text model, this simple approach achieves state-of-the-art (SOTA) results on MultiWOZ 2.2, and performance is on par with SOTA on MultiWOZ 2.1 and M2M. In addition, our analyses provide empirical results that contribute towards understanding how schema description augmentation can effectively constrain the model prediction.

\begin{figure*}[t!]
    \centering
    \includegraphics[width=\linewidth]{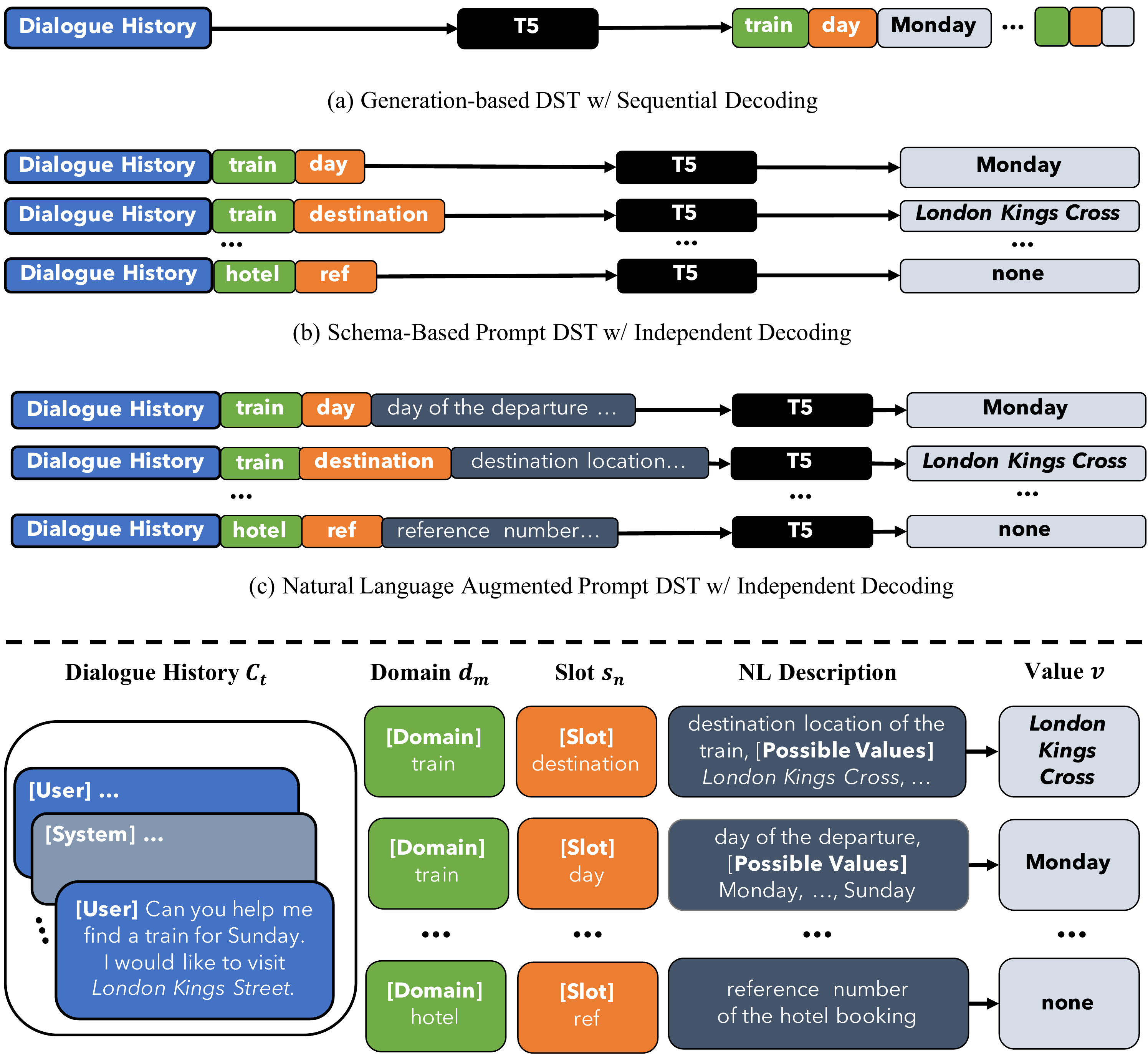}
    \caption{Overview of generative DST approaches for multi-domain scenario. The top three figures illustrate three different generative approaches considered in this paper and the bottom figure includes specific examples for dialogue history, domain names, slot names, natural language descriptions (types, set of valid values, etc.) for slots. Sub-figure (b)(c) demonstrate two prompt-based DST models proposed, where method in (c) includes additional natural language description of slots considered for tracking. Domain descriptions are omitted for brevity.}
    \label{fig:system}
\end{figure*}

\section{Related Work}
\subsection{Multi-Domain Dialogue State Tracking}
Task-oriented dialogue datasets \cite{shah2018building,henderson2014second}, have spurred the development of dialogue systems \cite{zhong2018global,chao2019bert}. Recently, to further examine the generalization abilities, large scale cross-domain datasets have been proposed \cite{budzianowski2018multiwoz,zang2020multiwoz,eric2019multiwoz,rastogi2020towards}. 
\textit{Classification-based} models \cite{ye2021slot,chen2020schema} pick the candidate from the oracle list of possible slot values. The assumption of the full access of the schema makes them have limited generalization abilities. On the other hand, \textit{generation-based} models \cite{wu2019transferable,kim2020efficient,lin-etal-2020-mintl} directly generate slot values token by token, making it possible to handle unseen domains and values. Most of these models require task-specific modular designs.

Recently, generation-based models that are built on large-scale autoregressive pretrained language models \cite{ham2020end,hosseini2020simple,peng2020soloist} achieve
promising state tracking results on MultiWOZ 2.0 and 2.1 when trained on additional supervision signals or dialogue corpus.
Both \citet{ham2020end} and \citet{hosseini2020simple} require dialogue acts as inputs. Both \citet{hosseini2020simple} and \citet{peng2020soloist} require DB search results as inputs. \citet{peng2020soloist} also leverages other dialogue corpora to finetune the language model. Our work requires only the dialogue state labels and does not utilize any external dialogue datasets. 


\subsection{Language Models}
Large-scale pretrained language models have obtained state-of-the-art performance on diverse generation and understanding tasks including bi-directional encoder style language models \cite{devlin2019bert,liu2019roberta}, auto-regressive language models \cite{radford2019language,brown2020language} and more flexible sequence-to-sequence language models \cite{raffel2020exploring}.
To adapt to dialogue tasks, variants of systems are finetuned on different dialogue corpora including chit-chat systems \cite{zhang2020dialogpt,adiwardana2020towards,roller2020recipes} and task-oriented dialogue systems \cite{mehri2019pretraining,wu2020tod,henderson2020convert,peng2020few}. We leave it as future work to leverage domain-adapted language models.

\subsection{Prompting Language Models}
Extending a language model’s knowledge via prompts is an active line of research. \citet{radford2019language} obtain empirical success by using prompts to guide zero shot generation without finetuning on any prompts. \citet{raffel2020exploring} uses task-specific prompts in both finetuning and testing phase. Recent studies have also tried to automatically discover prompts rather than writing them by humans \cite{jiang2020can}. Our proposed prompting method is largely inspired by this body of work. Instead of prompt engineering/generation, we focus on using available natural language descriptions of schema categories associated with database documentation as task-specific promptings for DST.



\section{Prompting Language Model for Dialogue State Tracking}
In this section, we first set up the notations that are used throughout paper, and then review the generative DST with the sequence-to-sequence framework.
Based on that, we formally introduce our prompt-based DST model and the corresponding backbone pretrained model.

\noindent
\textbf{Notation.}
For task-oriented dialogues considered in this paper,
a dialogue consists of a sequence of utterances alternating between two parties, $U_{1}, A_{1}, ..., U_{T}, A_{T}$,
where $U$ and $A$ represent the user utterance and the system response, respectively.
In a turn $t$, the user provides a new utterance \textit{$U_{t}$} and the system agent responds with utterance \textit{$A_{t}$}.
As shown in the bottom of \autoref{fig:system}, at turn $t$, we denote the dialogue context as $C_t=\{U_{1}, A_{1}, \ldots, A_{t-1}, U_{t}\}$, which excludes the latest system response $A_t$.
In this work, we assume a multi-domain scenario, in which case the schema contains $M$ domains $\mathcal{D}=\{d_1, \ldots, d_M\}$ and
$N$ slots $\mathcal{S}=\{s_1, \ldots, s_N\}$ to track as examples illustrated in \autoref{fig:system}.
$B_t$, the dialogue state at turn $t$, is then defined as a mapping from a pair ($d_m$, $s_n$) into values $v$.
Here, we define $B_t(d_m, s_n)=\phi$, if ($d_m$, $s_n$) is not in the current dialogue state.
In the given example of \autoref{fig:system}, the pair (domain=\textit{hotel}, slot=\textit{ref}) is not in the dialogue state, and the value ``\textit{none}'' is assigned.

\subsection{Generation-based DST with the Sequence-to-sequence Model}
There are primarily two decoding strategies for generation-based DST in the literature for inferring the dialogue state at a particular turn -- sequential (a) and independent (b)(c) -- both of which are explored in the paper as illustrated in \autoref{fig:system}. 

In the first case (top system (a) in \autoref{fig:system}), the dialogue history $C_t$ is taken as input to the encoder, and domain-slot-value triplets ($d_m$, $s_n$, $v$) are generated sequentially, where $B_t(d_m, s_n)\neq\phi$. This approach is adopted in many systems that leverage autoregressive LMs \cite{peng2020soloist,hosseini2020simple}. 
Despite being simple, this kind of sequential generation of multiple values is more likely to suffer from optimization issues with decoding long sequences resulting in lower performance. However, given its wide adoption in the literature, we still include this type of generative DST with the same backbone pretrained encoder-decoder Transformer model in our experiments.
To partially address this issue, \citet{lin2020mintl} propose a domain independent decoding where the decoder only have to generate a sequence of slot and value pairs within a specific given domain. Although their model leverages the same backbone model as ours, we empirically find that this form of strategy is still of limited effectiveness.

In the second case (middle two systems (b)(c) in \autoref{fig:system}), the values for each domain-slot pair are generated independently, potentially in parallel.  
The domain and slot names (embedded as continuous representations) are either the initial hidden state of the decoder \cite{kim2020efficient} or the first input of the decoder \cite{wu2019transferable}.
Values are either generated for all possible domain-slot $(d_m, s_n)$ pairs with a possible value of ``\textit{none}'' and/or there is a separate gating mechanism for domain-slot combinations not currently active.
Since we are interested in enriching the input with task-specific information, we focus on extending the independent decoding modeling for our prompt-based DST.

\subsection{Prompt-based DST}
In this section, we formally present the flow of our prompt-based DST with an encoder-decoder architecture.
Here, we are interested in an encoder-decoder model with a bi-directional encoder \cite{raffel2020exploring,lewis-etal-2020-bart}, in contrast with the uni-directional encoder used in autoregressive LMs \cite{radford2019language,gpt3}.

The input of the prompt-based DST is made up of a dialogue context $C_t$ and a task-specific prompt.
Here, we use two types of task-specific prompts, the domain-related prompt $X(d_m)$, and slot-related prompt $X(s_n)$, both of which are derived based on the given schema. We leave the discussion of two specific realizations of task-specific prompts to the later part of this section.
Specifically, all sub-sequences are concatenated with special segment tokens, i.e.,
``\texttt{[user]} $U_1$ \texttt{[system]} $A_1$ \ldots \texttt{[system]} $A_{t-1}$ \texttt{[user]} $U_t$ \texttt{[domain]} $X(d_m)$ \texttt{[slot]} $X(s_n)$'',
as input to the encoder, where \texttt{[user]}, \texttt{[system]}, \texttt{[domain]}, \texttt{[slot]} are special segment tokens for indicating the start of a specific user utterance, system utterance, domain-related prompt, and slot-related prompt, respectively.

Given this prompt-augmented input, the bi-directional encoder then outputs
\begin{equation}
    H_t = \text{Encoder}(C_t, X(d_m), X(s_n)),
\end{equation}
where $H_t\in\mathbb{R}^{L\times k}$ is the hidden states of the encoder, $L$ is the input sequence length, and $k$ is the encoder hidden size.
Then, the decoder attends to the encoder hidden states and decodes the corresponding slot value $B_t(d_m, s_n)$:
\begin{equation}
    B_t(d_m, s_n) = \text{Decoder}(H_t).
\end{equation}
The overall learning objective of this generation processing is maximizing the log-likelihood of $B_t(d_m, s_n)$ given $C_t$, $X(d_m)$ and $X(s_n)$, that is
\begin{equation}
\sum_{(m, n)} \log P(B_t(d_m, s_n)|C_t, X(d_m), X(s_n)).
\end{equation}
During inference, a greedy decoding procedure is directly applied, i.e., only the most likely token in the given model vocabulary is predicted at each decoding step.

\noindent
\textbf{Schema-Based Prompt.}
The first realization of task-specific prompt considered in this paper is based on the domain and slot names as defined in the task-dependent schema.
As shown in (b) of \autoref{fig:system}, given the domain name \textit{train} and the slot name \textit{day}, the specific prompt is in the form of
``\texttt{[domain]} \textit{train} \texttt{[slot]} \textit{day}''.
Different from \cite{lin-etal-2020-mintl,wu2019transferable} where the task-specific information is used in the decoder side, our symbol-based prompt as additional input to the bi-directional encoder can potentially achieve task-aware contextualizations.
Observing that users often revise/repair their earlier requests in dialogues, we posit that the resulting encoded representations can be more effectively used by the decoder for generating corresponding slot values.


\noindent
\textbf{Natural Language Augmented Prompt.}
One main drawback of symbol-based prompt is that those domain/slot names contain limited information that can be utilized by pretrained LMs.
In other words, those symbols from the custom schema are typically under-specified and unlikely to appear in corpus for LM pretraining.
Fortunately, documentation is commonly available for real-world databases, and it is a rich resource for domain knowledge
that allows dialogue systems to better understand the meanings of the abbreviated domain and slot names.
The documentation includes but is not limited to domain/slot descriptions and the list of possible values for categorical slots.
In this work, we experiment with a simple approach that augments the input by incorporating the domain description after the domain name and the slot description (with the sequence of values, if any) following the slot name, as illustrated in the system (c) in \autoref{fig:system}.

\subsection{Backbone Sequence-to-sequence Model}
Our prompt-based DST model is initialized with weights from a pretrained LM in an encoder-decoder fashion.
In this paper, we use the Text-to-Text Transformer (T5) \cite{raffel2020exploring} as our backbone model.
T5 is an encoder-decoder Transformer with relative position encodings \cite{shaw2018self}. We refer interested readers to the original paper for more details.

\section{Experiments}
\label{sec:experiment}

\subsection{Datasets}

\begin{table}[t]
    \small
    \centering
    \begin{tabular}{l@{\hskip3pt}c@{\hskip3pt}cc}
    \toprule
        Dataset &  MWOZ 2.2 &  MWOZ 2.1 & M2M  \\
        \midrule
        \# Domains & 8 & 8 & 2 \\
        \# Dialogues & 10438 & 10438 & 3008 \\
        \# Total Turns & 143004 & 143048  & 27120 \\
         Avg. Turns per Dial. & 13.70 & 13.70  & 9.01 \\   
         Avg. Toks per Turn & 13.23 & 13.18  & 8.28 \\   
        \# Cat. Slots & 21 & 0 & 0\\ 
        \# Non-Cat. Slots  & 40 & 37 & 12\\ 
        \midrule
        Domain Desc. & Y & N & N \\
        Slot Desc.   & Y & Y & N \\
        Value Set    & Y & N & N \\
        \bottomrule
    \end{tabular}
    \caption{Experiment data summary. The numbers are computed on all splits of the datasets. \texttt{MWOZ} stands for MultiWOZ. \texttt{Cat. Slots}\ and \texttt{Non-Cat. Slots}\ stand for categorical slots and non-categorical slots, respectively. The rows \texttt{Domain Desc.}\ and \texttt{Slot Desc.}\ indicate whether the corresponding dataset has natural language description for domains and slots, respectively. The row \texttt{Value Set} incates whether the corresponding dataset provides possible value set for categorical slots.
    \label{tab:data_sum}}
\end{table}

Table~\ref{tab:data_sum} summarizes the statistics of the datasets used in our experiments.

\noindent
\textbf{MultiWOZ}
\cite{budzianowski2018multiwoz} is a multi-domain task-oriented dialogue dataset that contains over 10K dialogues across 8 domains.
It is a collection of human-human written conversations and has been one of the most popular benchmarks in the DST literature.
Since its initial release, many erroneous annotations and user utterances have been identified and fixed in subsequent versions, i.e., 
MultiWOZ 2.1 \cite{eric2019multiwoz}
and MultiWOZ 2.2 \cite{zang2020multiwoz}.
In addition, MultiWOZ 2.1 provides 2-3 descriptions for every slot in the dataset. We randomly sample one of them and use the same descriptions for every experiment. The original dataset does not have domain descriptions and possible values so these are omitted in the corresponding experiments.
MultiWOZ 2.2 further provides descriptions of domain and slot as well as possible values for categorical slots. 

\noindent
\textbf{Machines Talking To Machines
(M2M)} \cite{shah2018building}
is a framework that combines simulation and online crowdsourcing. Templates of each dialogue are first generated and then online workers rewrite the conversations to make them human-readable while preserving the meaning. It provides 3,000 dialogues spanning 2 domains. 
The restaurant domain is denoted as \texttt{Sim-R} and the movie domain is denoted as \texttt{Sim-M}. 
Since there are no descriptions provided in the corpus, we take existing descriptions from other corpora that have the same slots.
Specifically, descriptions for the restaurant domain are taken from MultiWOZ 2.2,
whereas descriptions for the movie domain are taken from SGD \cite{rastogi2020towards}.
All slots in M2M are covered. Since all slots are non-categorical, the descriptions do not include the possible values.






\noindent{\bf Evaluation Metric}.
The standard joint goal accuracy (JGA) is used as the evaluation metric.
It treats a prediction as correct only if for every domain all slots exactly match the ground-truth values.
For MultiWOZ 2.1 and 2.2, we use the official evaluation script from the DSTC8 challenge \cite{rastogi2020schema}.\footnote{\url{https://github.com/google-research/google-research/tree/master/schema_guided_dst\#evaluation-on-multiwoz-21}}
For M2M, we adopt the above evaluation scripts with simple modifications.

\subsection{MultiWOZ 2.2: Fully Annotated Natural Language Augmented Prompt}
We present the evaluation results on MultiWOZ 2.2 in \autoref{tab:woz2.2}.
The following baseline models are considered: TRADE \cite{wu2019transferable}, DS-DST \cite{zhang2019find} and Seq2Seq-DU \cite{feng2020sequence}.
Similar to ours, the decoding strategy of TRADE is independent. 
However, the sum of domain and slot embeddings are the first input of the decoder, which makes their dialogue history representation not task-aware contextualized. 
The sequential decoding strategy is worse than the independent decoding strategy by over 5\% with both T5-small and T5-base. 
Even with T5-small (almost half the model size of BERT-base which is used in most previous benchmark models), our system achieves the SOTA performance using the independent decoding.
As expected, T5-base systems outperform T5-small systems. 
With the augmentation of descriptions, we improve the overall JGA by over 1\% in both T5-small and T5-base.  

\begin{table}[t]
    \small
    \centering
    \begin{tabular}{lcc}
    \toprule
        \textbf{Models}  & \textbf{Pretrained-Model/ \# Para.}   & \textbf{JGA} \\
        \midrule
        TRADE & N  & 48.6   \\
        DS-DST & BERT-base / (110M) & 51.7 \\
        Seq2Seq-DU & BERT-base / (110M)  & 54.4 \\
        \midrule
        \midrule
        Sequential & T5-small / (60M)  & 48.9\\
        Sequential  & T5-base / (220M)  & 51.2  \\
        \midrule
        Independent  &  T5-small / (60M) & 55.2 \\
        \quad \textit{w.} desc &  T5-small / (60M) & 56.3  \\
        Independent &  T5-base / (220M) & 56.7 \\
        \quad \textbf{\textit{w.} desc} &  T5-base / (220M)  & \textbf{57.6}  \\
        \bottomrule
    \end{tabular}
    \caption{Results on MultiWOZ 2.2. All numbers are reported in joint goal accuracy (JGA)(\%). 
    \textit{w.} desc means the model is trained with the description.  \# Para. stands for the number of model parameters.}
    \label{tab:woz2.2}
\end{table}

\subsection{MultiWOZ 2.1: Partially Annotated Natural Language Augmented Prompt}
Different from MultiWOZ 2.2 studied in the previous section, MultiWOZ 2.1 only contains natural language descriptions for slots but not domains.
In addition, there is no possible slot value information.

The evaluation results on MultiWOZ 2.1 are shown in Table~\ref{tab:woz2.1}, where we compare with 
TRADE \cite{wu2019transferable}, MinTL \cite{lin-etal-2020-mintl}, SST \cite{chen2020schema}, TripPy \cite{heck2020trippy}, Simple-TOD \cite{hosseini2020simple}, SOLOIST \cite{peng2020soloist} and TripPy+SCORE \cite{yu2020score}.
Note that both SOLOIST and TripPY+SCORE use external dialogue datasets to finetune their models.

As expected, we observe that T5-base models perform consistently better than T5-small models.
Moreover, using descriptions consistently improves the performance of both models.
All our models outperform baselines that do not use extra dialogue data. 
It is worth noting that comparing with MinTL (T5-small), our model is better by over 4\% even without descriptions. 
Further, our T5-small system is even better than MinTL built on BART-LARGE \cite{lewis-etal-2020-bart} which has substantially more parameters.
Similar to ours, MinTL leverages a sequence-to-sequence LM.
One difference is that their domain information is fed only to the decoder while our approaches enables task-aware contextualization by prompting the LMs
with domain and slot information on the encoder side.
Another difference is that they jointly learn DST together with dialogue response generation, which provides more supervision signals.
Therefore, the better performance of our systems implies that schema-driven prompting is effective. 

Lastly, compared with MultiWOZ 2.2, the performance gain brought by augmenting natural language descriptions is less pronounced which is likely caused by the reduced information available in MultiWOZ 2.1 descriptions.

\begin{table}[t]
    \small
    \centering

    \begin{tabular}{l@{\hskip3pt}c@{\hskip3pt}c}
    \toprule
        \textbf{Models} & \textbf{Pretrained-Model / \# Para.} &  \textbf{JGA}  \\
        \midrule
        TRADE & N & 45.60\\
        MinTL  & T5-small / (60M) & 50.95\\
        MinTL   & BART-large / (406M) & 53.62 \\
        SST  & N &  55.23 \\
        TripPy   & BERT-base / (110M) &  55.29 \\
        Simple-TOD\footnote{We report the results of no label-cleaning here to allow fair comparisons.}  & GPT2 / (117M) & 55.72  \\
        \midrule
        *SOLOIST  &  GPT-2 / (117M) & 56.85  \\
        *\textbf{TripPy + SCORE} & ROBERTA-large / (355M) & \textbf{60.48} \\
        \midrule
        \midrule 
        Independent & T5-small / (60M)   & 55.37  \\
        \quad \textit{w.} desc  &  T5-small / (60M)  & 56.12  \\
        Independent & T5-base / (220M)  &  56.39 \\
        \quad \textbf{\textit{w.} desc}  & T5-base / (220M) &  \textbf{56.66}\\
        \bottomrule
    \end{tabular}
        \caption{Results on MultiWOZ 2.1. All numbers are reported in joint goal accuracy (JGA)(\%). \textit{w.} desc means the model is trained with the description. * means extra dialogue data is used to finetune the language model. \# Para. stands for the number of model parameters. } 
    \label{tab:woz2.1}
\end{table}

\subsection{M2M: Borrowed Natural Language Augmented Prompt}
Table~\ref{tab:m2m} shows the evaluation results on M2M. 
In this case, all natural language descriptions are directly borrowed from dialogue datasets that are annotated in a different manner.
We achieve the SOTA performance on Sim-R and Sim-M+R while being comparable on Sim-M.  
The improvements of descriptions are only evident on the restaurant domain. 
The lack of improvement from slot descriptions for the movie domain may be because the slot descriptions do not add much beyond the slot name (compared to "category" for the restaurant domain) or that it has slots that generalize better across domains (e.g. date, time, number of people).

\begin{table}[t]
    \small
    \centering

    \begin{tabular}{lccc}
    \toprule
        \textbf{Models}  &  \textbf{Sim-M} & \textbf{Sim-R} & \textbf{Sim-M+R} \\
        \midrule
        \cite{rastogi2017scalable} & 96.8 & 94.4 & --  \\
        \midrule
         \cite{rastogi2018multi} & 50.4 & 87.1  & 73.8 \\
         \cite{chao2019bert} & 80.1 & 89.6 & --  \\
         \cite{heck2020trippy} & \textbf{83.5} & 90.0 & -- \\
         \midrule
        Independent &  83.3 & 89.6 &  \textbf{88.0}  \\
        \quad \textit{w.} desc & 81.0 & \textbf{90.6} & 86.4  \\
        \bottomrule
    \end{tabular}
        \caption{Results on M2M. All numbers are reported in joint goal accuracy(JGA)(\%). \cite{rastogi2017scalable} should be considered as a kind of oracle upper bound performance because the target slot value is guaranteed to be in the candidate list and consider by the model.}
    \label{tab:m2m}
\end{table}

\section{Analysis}
\subsection{Breakdown Evaluation for MultiWOZ}
In \autoref{tab:cat}, we follow the categorization provided in \cite{zang2020multiwoz} and show the breakdown evaluation of categorical and non-categorical slots on MultiWOZ 2.2. 
As we can see, the breakdown accuracy scores for both categorical and non-categorical slots are pretty consistent with the overall JGA.
For both T5-small and T5-base models, models with sequential decoding perform worse than the corresponding models with independent decoding for both categorical and non-categorical slots.
In particular, the independent decoding models achieve more pronounced improvement in categorical slots indicating that the task-specific prompt is very helpful for guiding the decoder to predict valid values. 
When comparing models using natural language description with those not, we observe performance gains for both types of slots for T-base but only non-categorical slots for T5-small. It is likely that the smaller size of T5 has limited representation capability to effectively utilize the additional textual description information regarding types and possible values. 


\begin{table}[t]
    \small
    \centering

    \begin{tabular}{lc@{\hskip3pt}c@{\hskip3pt}c}
    \toprule
        \textbf{Models}  &  \textbf{JGA} & \textbf{CAT} & \textbf{NON-CAT} \\
        \midrule
        Sequential (T5-small)  & 48.9 & 61.3 & 69.0  \\
        Sequential (T5-base) & 51.2 & 62.9 & 70.9  \\
        \midrule
        Independent (T5-small)  & 55.2 & 71.4 & 75.2  \\
        \quad \textit{w.} desc & 56.3 & 71.1 & 76.2  \\
        \quad \textit{w. only} slot desc & 55.2 & 70.4 & 75.8  \\
        \quad \textit{w. only} domain desc & 54.3 & 70.1 & 75.4  \\
        \quad \textit{w. only} slot + domain desc & 55.9 & 71.2 & 76  \\
        \midrule
        Independent (T5-base)  & 56.7 & 71.6 &  76.3 \\
        \quad \textit{w.} desc & 57.6 & 72.4 & 76.8 \\
        \bottomrule
    \end{tabular}
        \caption{Slot type breakdown results on the test set of MultiWOZ 2.2. All numbers are reported in joint goal accuracy(JGA) (\%). CAT and NON-CAT correspond to categorical slots JGA and non-categorical slots JGA, respectively. \textit{w.} desc indicates that the model is trained with the full description.}
    \label{tab:cat}
\end{table}

\vspace{-3mm}

\subsection{Ablation Study on Schema Descriptions}
To understand what parts of the schema descriptions are most important, we experiment with three kinds of description combinations on MultiWOZ 2.2 using the T5-small configuration: \textbf{(i)} excludes the list of possible values for categorical slots \textbf{(ii)} excludes slot descriptions \textbf{(iii)} excludes domain descriptions. For \textbf{(i)}, there is an 0.4\% point drop in JGA, validating that value sets can successfully constrain the model output, as we illustrate in \autoref{tab:correct}. For  \textbf{(ii)}, there is a 0.8\% point drop in JGA. And for \textbf{(iii)}, there is a 0.1\% point drop in JGA. This shows that slot descriptions are the most important part of the schema prompts and domain descriptions are relatively less effective. This is probably due to the fact that there are 61 slots in MultiWOZ 2.2 but only 8 domains. Also, the domain names are all self-contained single words.

\subsection{The Effectiveness of Natural Language Augmented Prompt}
In order to understand the benefit of natural language augmented prompt, we focus on analyzing the examples where the description augmented model correctly tracks the dialogue state while the unaugmented one fails.
Based on our analysis of T5-base model on MultiWOZ 2.2, the most common errors are either misses of gold slots or
over-predictions of irrelevant slots (82.8\% of all errors). The remaining error cases are correct slot predictions with wrong slot values (17.2\%). 

\begin{table*}[t]
    \small
    \centering

    \vspace{-0.5\baselineskip}
    \begin{tabular}{lp{12.0cm}}
        \toprule
        \textbf{Database \textit{Train}} & \textbf{Slot Descriptions || Possible Values}\\
        \textit{arriveby}   & \textbf{arrival time of the train} \\
        \textit{destination}  & destination of the train \textbf{||} Birmingham New Street, \textbf{London Kings Cross}, ..., Stevenage\\
        
        \midrule
        \textbf{Dialogue History}  & ... \textit{[SYS]} The earliest being 19:09 and arriving by 20:54. Would that work for you?\\ & \textit{[USR]} Yes, I think the \textbf{20:54 arrival time} should work.\\
        \textbf{no desc.} & \textit{(train, day, friday) (train, departure, leicester) (train, destination, cambridge) (train, leaveat, 19:00)} \\
        \textbf{desc.} & \textit{\textbf{(train, arriveby, 20:54)} (train, day, friday) (train, departure, leicester) (train, destination, cambridge) (train, leaveat, 19:00)} \\
         \midrule
        \textbf{Dialogue History}  & \textit{[USER]} I need to find a train going to Leicester that arrives by \textbf{4:45 PM}. Do you know of one?\\
        \textbf{no desc.} & \textit{(train, arriveby, 04:45) (train, destination, leicester)} \\
        \textbf{desc.} & \textit{\textbf{(train, arriveby, 16:45)} (train, destination, leicester)}\\
        \midrule
        \textbf{Dialogue History}  &  \textit{[USER]} Can you help me find a train for Sunday. I would like to visit \textbf{London Kings Street}.\\
        \textbf{no desc.} &  \textit{(train, destination, London Kings Street) (train, day, Sunday)} \\
        \textbf{desc.} & \textit{\textbf{(train, destination, London Kings Cross)} (train, day, Sunday)}\\
        \bottomrule
    \end{tabular}
        \caption{
        Examples for \texttt{train} domain dialogues where the description-augmented (``desc.'') model make the correct state predictions but the unaugmented models (``no desc.'') fails. The correctly predicted triplets are in bold.
    }
    \label{tab:correct}
\end{table*}

\begin{table*}[t!]
    \small
    \centering

    \vspace{-0.5\baselineskip}
    \begin{tabular}{lp{12.0cm}}
        \toprule
        \multicolumn{2}{c}{\textbf{53.33\%: Annotation Errors}} \\
        \textbf{Dialogue History}  &  ...\textit{[SYSTEM]} Out of the 21 restaurant choices, one is the \textbf{Yippee Noodle Bar which is moderately priced in the centre of town}. Would you like to make a reservation?\\ 
         &  \textit{[USER]} \textbf{That sounds great}, what is the postcode? \\
        \textbf{Gold} &  \textit{()}\\
        \textbf{desc. Prediction} & \textit{(restaurant, area, centre) (restaurant, pricerange, moderate) (restaurant, name, yippee noodle bar}\\
         \midrule
        \multicolumn{2}{c}{\textbf{20.00\%: Unable to Capture System Information}} \\
        \textbf{Dialogue History}  & ... \textit{[SYSTEM]} There is TR6679. \textbf{It leaves at 19:35 and arrives at 19:52}. Is that good for you?\\
        &  \textit{[USER]} \textbf{Sounds good}. May I have the travel time and ticket price, please? \\
        \textbf{Gold} & \textit{(train, arriveby, 19:52) (train, leaveat, 19:35)} \\
        \textbf{desc. Prediction} & \textit{()} \\
         \midrule
        \multicolumn{2}{c}{\textbf{16.66\%: Unable to Mention Slot Provided by User}} \\
        \textbf{Dialogue History}  & ... \textit{[USER]} Do you happen to know if there is a nightclub in the centre?\\ 
        & \textit{[SYSTEM]} Yes, we have FIVE nightclubs in the centre of town. Is there a particular one you're looking for?\\
        &  \textit{[USER]} \textbf{I don't care which one you recommend}, but can you tell me the entrance fee and address?\\ 
        \textbf{Gold} & \textit{(attraction, area, centre)  (attraction, type, nightclub)} \textbf{(attraction, name, dontcare)} \\
        \textbf{desc. Prediction} & \textit{(attraction, area, centre) (attraction, type, nightclub)} \\
         \midrule
        \multicolumn{2}{c}{\textbf{10.00\%: Incorrect Value Reference}} \\
        \textbf{Dialogue History}  & \textit{[USER]} Hi can you help me find a very nice Italian restaurant near the centre of cambridge? \\ 
        & \textit{[SYSTEM]} Please specify your price range.\\
        & \textit{[USER]} \textbf{It does not matter}. \\
        \textbf{Gold} & \textit{(restaurant, area, centre) (restaurant, food, italian) \textbf{(restaurant, pricerange, dontcare)}} \\
        \textbf{desc. Prediction} & \textit{(restaurant, area, centre) (restaurant, food, italian) \textbf{(restaurant, pricerange, expensive)}} \\
        \bottomrule
    \end{tabular}
        \caption{The most common error types of our best model(t5-base \textit{w/ desc.}) and corresponding examples.}
    \label{tab:error}
\end{table*}

We provide representative examples for which the description augmented system correctly tracks the dialogue states but not the unaugmented one in \autoref{tab:correct}. In the first example, the phrases in the dialogue history are partially matched to the slot description of \textit{arriveby} making it easier for the description-augmented system to detect the mention of the correct slot. For the second example, the type information in the description implicitly guides the model to focus on time-related information leading the correct output of the normalized time expression, 16:45. In contrast, the model without descriptions only generates the partial answer 4:45, ignoring PM. Lastly, "London Kings Street" is a typographical error in this case. By utilizing the provided possible values included in the slot descriptions, the model is able to generate the correct slot value without spelling error, demonstrating that the natural language augmented prompt can successfully constrain the model output and potentially provides robustness to the dialogue state tracking system.

\vspace{-3mm}

\subsection{Error Analysis of Natural Language Augmented Prompt-based DST}
Here, we further carry out error analyses into the natural language augmented prompt-based T5-base model on MultiWOZ 2.2. As shown in \autoref{tab:error}, we randomly sample 50 turns and categorize them into different types. In summary, there are four types of errors:
\textbf{(i)} The most common error type is annotation error in which the model prediction is actually correct, which is similar to the findings of \cite{zhou2019multi}. \textbf{(ii)} 20\% of the errors come from model failing to capture information provided by the system.\footnote{There is label inconsistency in the MultiWoZ as pointed out by \cite{zhou2019multi}. If the user confirms the booking or gives a positive response, then the dialogue states in the previous system utterance should be grounded. However, this rule is not always followed in the dataset construction. So to some extent, this type of error is inevitable.} \textbf{(iii)} 16.66\% of the errors are caused by the model misses of at least one gold slot. \textbf{(iv)} 10\% of the errors are correct slot predictions with the wrong corresponding values. In general, most errors are likely caused by the lack of explicit modeling of user-system interactions.

\section{Conclusion}
In this work, we propose a simple but effective task-oriented dialogue system based on large-scale pretrained LM. We show that, by reformulating the dialogue state tracking task as prompting knowledge from LM, our model can benefit from the knowledge-rich sequence to sequence T5 model. Based on our experiments, the proposed natural language augmented prompt-based DST model achieve SOTA on MultiWOZ 2.2 and comparable performance on MultiWOZ 2.1 and M2M to recent SOTA models. Moreover, our analyses provide evidence that the natural language prompt is effectively utilized to constrain the model prediction.

\section*{Acknowledgements}
This research was supported in part by a grant from Allstate. We would like to thank the reviewers for their constructive feedback.
\bibliography{custom}
\bibliographystyle{acl_natbib}

\clearpage
\renewcommand{\appendixpagename}{Supplementary Material}
\appendix
\appendixpage

\section{Implementation Details}
The backbone models we use for finetuning are T5-small(~60M  parameters) and T5-base(~220M parameters). We use the pretrained checkpoint from \textit{transformers}  library\footnote{\url{https://huggingface.co/t5-small}
,\url{https://huggingface.co/t5-base}}. For T5-small, we train the model with a batch size 4, a learning rate of 5e-5 for 3 epochs. For T4-base, we train the model with a batch size of 64, a learning rate of $5\mathrm{e}{-4}$ for 2 epochs. Both models are trained using Adam\cite{loshchilov2018decoupled}. We don't use any text or label normalization scripts like \cite{wu2019transferable,hosseini2020simple}.\\

For MultiWOZ 2.1 and 2.2, following many previous works\cite{wu2019transferable}, since \textit{police} and \textit{hospital} domains only appear in the training set, we exclude them in all our experiments.

\section{Descriptions}
We show the descriptions of  M2M and MultiWOZ 2.1in Table\ref{tab:m2mDESC} and Table\ref{tab:2.1DESC}
\label{sec:appendix}
\begin{table*}[t]
    \small
    \centering
    \caption{Domain and slot descriptions of M2M used in our experiments. The descriptions of the movie domain is taken from \cite{rastogi2020schema} and the descriptions of the restaurant domain is taken from \cite{zang2020multiwoz}.}
    \label{tab:m2mDESC}
    \begin{tabular}{cccc}
    \toprule
        \multicolumn{4}{c}{Sim-M}\\
        \midrule
        \textbf{Domain}  &  \textbf{Domain Description} & \textbf{Slot} & \textbf{Slot Description} \\
        \midrule
        Movie & A go-to provider for finding movies,  & theatre\_name  & the name of the theatre where the movie is playing \\
         & searching for show times and booking tickets  & movie  &  name of the movie  \\
         &  & date  & date of the show booking \\
         &  & time  & time of the show booking \\
         &  & num\_people & number of people to purchase tickets for \\
        \bottomrule
        \toprule
        \multicolumn{4}{c}{Sim-R}\\
        \midrule
         \textbf{Domain}   & \textbf{Domain Description} & \textbf{Slot}  & \textbf{Slot Description} \\
        Restaurant & find places to dine and whet your appetite & price\_range & price budget for the restaurant \\
        & & location & the location or area of the restaurant \\
        & & restaurant\_name & the name of the restaurant \\
        & & category & the cuisine of the restaurant you are looking for \\
        & & num\_people & how many people for the restaurant reservation\\
        & & date & date of the restaurant booking \\
        & & time & time of the restaurant booking \\
        \bottomrule
    \end{tabular}
\end{table*}

\begin{table*}[t]
    \small
    \centering
    \caption{The randomly sampled descriptions of MultiWOZ 2.1 used in all our experiments.}
    \label{tab:2.1DESC}
    \begin{tabular}{ccc}
    \toprule
        \multicolumn{3}{c}{MultiWOZ 2.1}\\
        \midrule
         \textbf{Domain}   & \textbf{Slot} & \textbf{Slot Description} \\
         taxi & leaveat & what time you want the taxi to leave your departure location by\\
         taxi & destination & destination of taxi\\
         taxi  & departure & what place do you want to meet the taxi\\
         taxi   & arriveby & when you want the taxi to drop you off at your destination\\
         restaurant  & book people & number of people booking the restaurant\\
         restaurant  & book day &  what day of the week to book the table at the restaurant\\
        restaurant  & book time & time of the restaurant booking\\
        restaurant  & food & food type for the restaurant\\
        restaurant  & pricerange &  price budget for the restaurant\\
        restaurant  & name & name of the restaurant\\
          restaurant  & area & preferred location of restaurant\\
          train & destination & destination of the train\\
          train  & day &  what day you want to take the train\\
        train  & departure &  departure location of the train\\
        train  & arriveby & what time you want the train to arrive at your destination station by\\
        train  & book people  &  number of people booking for train\\
        train  & leaveat &  when you want to arrive at your destination by train\\
        hotel  & pricerange  & preferred cost of the hotel\\
        hotel  & type &   type of hotel building\\
        hotel  & parking  &  parking facility at the hotel\\
        hotel  & book stay &    length of stay at the hotel\\
        hotel  & book day  &     day of the hotel booking\\
        hotel & book people &      how many people are staying at the hotel\\
        hotel  &   area  &  rough location of the hotel\\
        hotel  & stars  &     rating of the hotel out of five stars\\
        hotel  &  internet  &  whether the hotel has internet\\
        hotel  &  name   &   which hotel are you looking for\\
        attraction   & type &   type of attraction or point of interest\\
        attraction &  area    &    area or place of the attraction\\
        attraction  &  name   &  name of the attraction\\
        \bottomrule
    \end{tabular}
\end{table*}

\end{document}